\documentclass[10pt,twocolumn,letterpaper]{article}

\usepackage{iccv}
\usepackage{times}
\usepackage{epsfig}
\usepackage{graphicx}
\usepackage{amsmath}
\usepackage{amssymb}
\usepackage{wasysym}
\usepackage[accsupp]{axessibility}
\usepackage{xcolor}
\usepackage{color, colortbl}
\definecolor{citecolor}{HTML}{2980b9}
\definecolor{linkcolor}{HTML}{c0392b}
\definecolor{notecolor}{HTML}{6495ED}
\definecolor{notecolor2}{HTML}{4682B4}
\usepackage[hidelinks,pagebackref=true,breaklinks=true,colorlinks,bookmarks=false,citecolor=citecolor,letterpaper=true,linkcolor=linkcolor]{hyperref}
\usepackage[utf8]{inputenc}
\usepackage[small]{caption}
\usepackage{amsthm}
\usepackage{booktabs}
\usepackage{algorithm}
\usepackage{algorithmic}
\usepackage{adjustbox}
\usepackage{makecell}
\usepackage{multirow}
\usepackage{marvosym}

\makeatletter
  \newcommand\figcaption{\def\@captype{figure}\caption}
  \newcommand\tabcaption{\def\@captype{table}\caption}
\makeatother
\newcommand\blfootnote[1]{%
  \begingroup
  \renewcommand\thefootnote{}\footnote{#1}%
  \addtocounter{footnote}{-1}%
  \endgroup
}
\iccvfinalcopy % *** Uncomment this line for the final submission

\def\iccvPaperID{1521} % *** Enter the ICCV Paper ID here

\ificcvfinal\fi

\begin{document}

%%%%%%%%% TITLE
\title{\vspace{-0.4cm}ViewRefer: Grasp the Multi-view Knowledge for 3D Visual Grounding\\with GPT and Prototype Guidance\vspace{-0.4cm}}
% \\by Linguistic-dense GPT and Prototypes

\author{
\ \ \ \ \ \ \ \ \ \ \ \ \ \ \ \ \ \ \ \ \ \ \ \ \ \ \ \ \ 
\and Zoey Guo\textsuperscript{1,2*}
\and
Yiwen Tang\textsuperscript{1*}
% Shanghai Artificial Intelligence Laboratory\\
\and
Ray Zhang\textsuperscript{1,2*}\ \ \ \ \ \ \ \ \ \ \ \ \ \ \ \ \ \ \ \ \ \ \ \ \ \ \vspace{-0.1cm}
\and
Dong Wang\textsuperscript{1\Letter}
\and
Zhigang Wang\textsuperscript{1}
\and
Bin Zhao\textsuperscript{1, 3\Letter}
\and
Xuelong Li\textsuperscript{1, 3}
\vspace{0.1cm}
\and
\textsuperscript{\rm 1} Shanghai Artificial Intelligence Laboratory
\and
\textsuperscript{\rm 2} The Chinese University of Hong Kong
\and
\ \ \ \ \ \ \ \ \ \ \ \ \ \ \ \ \ \ \ \ \ \ \ \ \textsuperscript{\rm 3} Northwestern Polytechnical University\ \ \ \ \ \ \ \ \ \ \ \ \ \ \ \ \ \ \ \
\and
{\tt\small \{tangyiwen, wangdong, zhaobin\}@pjlab.org.cn}
}

\maketitle
\blfootnote{*\ Equal Contribution.}
\blfootnote{\Letter\  Corresponding authors}
\ificcvfinal\thispagestyle{empty}\fi

%%%%%%%%% ABSTRACT
\begin{abstract}
Understanding 3D scenes from multi-view inputs has been proven to alleviate the view discrepancy issue in 3D visual grounding. However, existing methods normally neglect the view cues embedded in the text modality and fail to weigh the relative importance of different views.
In this paper, we propose \textbf{ViewRefer}, a multi-view framework for 3D visual grounding exploring how to grasp the view knowledge from both text and 3D modalities. 
For the text branch, ViewRefer leverages the diverse linguistic knowledge of large-scale language models, e.g., GPT, to expand a single grounding text to multiple geometry-consistent descriptions.
Meanwhile, in the 3D modality, a transformer fusion module with inter-view attention is introduced to boost the interaction of objects across views. 
On top of that, we further present a set of learnable multi-view prototypes, which memorize scene-agnostic knowledge for different views, and enhance the framework from two perspectives: a view-guided attention module for more robust text features, and a view-guided scoring strategy during the final prediction.
With our designed paradigm, ViewRefer achieves superior performance on three benchmarks and surpasses the second-best by \textbf{+2.8\%}, \textbf{+1.5\%}, and \textbf{+1.35\%} on Sr3D, Nr3D, and ScanRefer. Code is released at \url{https://github.com/Ivan-Tang-3D/ViewRefer3D}.
\end{abstract}

\begin{figure}[t!]
\vspace{0.2cm}
    \includegraphics[width=\linewidth]{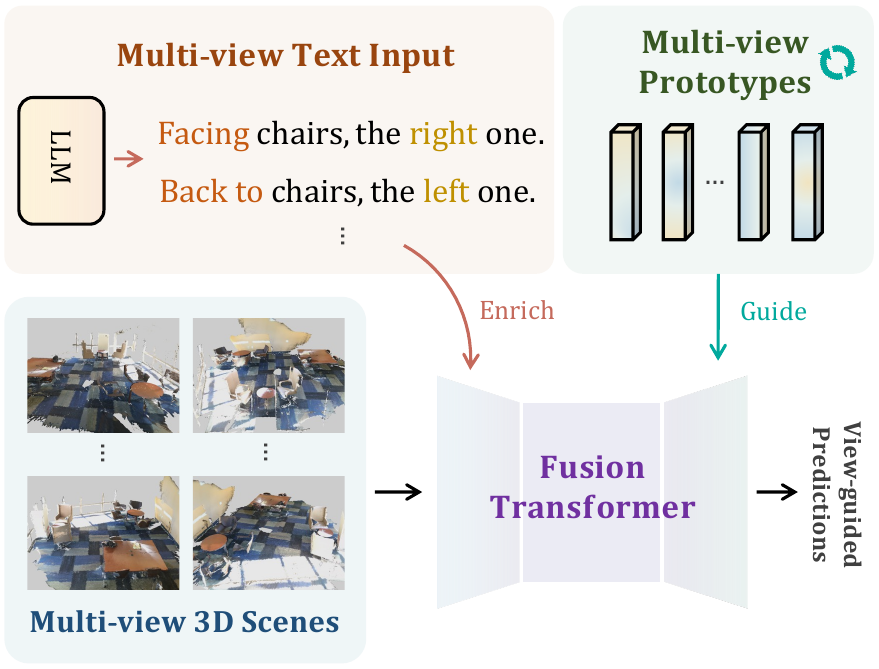}
  \caption{\textbf{The Paradigm of ViewRefer.} With our proposed LLM-expanded texts and multi-view prototypes, ViewRefer adopts a fusion transformer to effectively grasp view knowledge from multi-modal data, which achieves superior 3D grounding performance.}
  \label{teaser}
  \vspace{-0.1cm}
\end{figure}

\vspace{-0.3cm}
\section{Introduction}
\vspace{-0.1cm}
The aim of visual grounding is to ascertain the precise location of an object from an image or a 3D scene according to given query texts. The field of 2D visual grounding~\cite{kamath2021mdetr,yang2019fast,liu2019learning} has undergone significant advances recently.
Meanwhile, the developments in embodied agent~\cite{savva2019habitat,duan2022survey}, vision-language navigation~\cite{zhu2020vision,wang2019reinforced}, and autonomous driving~\cite{hausler2021patch} have also stimulated increasing attention on 3D visual grounding~\cite{achlioptas2020referit3d,chen2020scanrefer}.%,he2021transrefer3d,huang2022multi,roh2022languagerefer,yuan2021instancerefer,zhao20213dvg}. 

Inspired by the strategies in 2D counterparts, most 3D visual grounding methods adopt a two-stage pipeline, which first detect all object proposals in the scene and then ground the target ones. 
Unlike 2D images with fixed object positions, the large-scale 3D scenes consist of irregular-distributed point clouds with intricate spatial information, which is view-invariant and causes more challenges. As discussed in previous works~\cite{roh2022languagerefer,huang2022multi}, one urging issue is \textit{view discrepancy}, caused by the uncertain perspective between the intelligent agent (model) and the commander (grounding text giver). Given relative positions between objects, the text descriptions are supposed to change according to different viewpoints, e.g., when turning the view from ``facing" into ``back to'', the ``right'' chair should be rectified as a ``left" one. Unfortunately, the public available datasets~\cite{achlioptas2020referit3d,chen2020scanrefer} for 3D visual grounding only provide one text query corresponding to point clouds of uncertain viewpoints.

% For the view discrepancy issue, LanguageRefer~\cite{roh2022languagerefer} manually corrects the view of training data to the speaker's view, but shows limited  performance on unseen test data.
% MVT~\cite{huang2022multi} attempts to address this issue by proposing a preliminary solution that aggregates 3D information from different perspectives to create a view-robust representation. 

To alleviate such potential misalignment, existing methods either manually align the 3D scenes to the paired texts~\cite{roh2022languagerefer}, or simultaneously feed multiple views into the network for better view robustness~\cite{huang2022multi}.
However, these methods have two major limitations. 
Firstly, they only focus on solving the view dependence issue from the 3D modality, while neglecting the lack of view cues within text input.  
Secondly, for the multi-view input, they introduce no specifically designed modules to capture the view knowledge, which is yet significant to discriminate the relative importance of each view. Therefore, we ask: \textit{Can we explicitly grasp the view knowledge from both text and 3D modalities to further boost the 3D grounding performance?}

To this end, we propose \textbf{ViewRefer}, a multi-view framework for 3D visual grounding, which captures sufficient view cues from both text and 3D modalities to understand the spatial inter-object relation. Our overall paradigm is shown in Figure~\ref{teaser}.
For the text modality, we leverage large-scale language models (LLMs) to expand the input grounding text with view-related descriptions. Such LLM-expanded texts can capture sufficient multi-view semantics inherited from LLMs' linguistic knowledge and perform better grounding performance for the target objects. For the 3D modality, we take as input the multi-view 3D scenes and adopt a fusion transformer for 3D-text feature interactions. In each block, we apply different attention mechanisms to exchange information across modalities, views, and objects, contributing to thorough multi-modal fusion.
On top of that, we further introduce a set of learnable multi-view prototypes, which aims to capture the inter-view knowledge during training. The guidance of prototypes lies in two aspects. The first complements input text features with adaptive multi-view semantics, the second refines the final output by weighing the importance of different views. Both of them provide high-level guidance for multi-view visual grounding in ViewRefer.

To demonstrate the effectiveness of our approach, we evaluate its performance on three commonly used benchmarks, i.e., Sr3D~\cite{achlioptas2020referit3d}, Nr3D~\cite{achlioptas2020referit3d} and ScanRefer~\cite{chen2020scanrefer}, where ViewRefer consistently achieves superior performance, surpassing the second-best by \textbf{+2.8\%}, \textbf{+1.5\%}, \textbf{+1.35\%}, respectively. The main contributions of our paper are summarized as follows:
\begin{itemize}
    \item We propose ViewRefer, a multi-view framework for 3D visual grounding, which grasps view knowledge to alleviate the challenging view discrepancy issue.
    
    \item For the text and 3D modalities, we respectively introduce LLM-expanded grounding texts and a fusion transformer for capturing multi-view information.

    \item We present multi-view prototypes to provide high-level guidance to our framework, which contributes to superior 3D grounding performance.
    
\end{itemize}

\section{Related Work}
\paragraph{3D Visual Grounding.}

\begin{figure*}[t!]
\vspace{0.2cm}
\centering
\includegraphics[width=\textwidth]{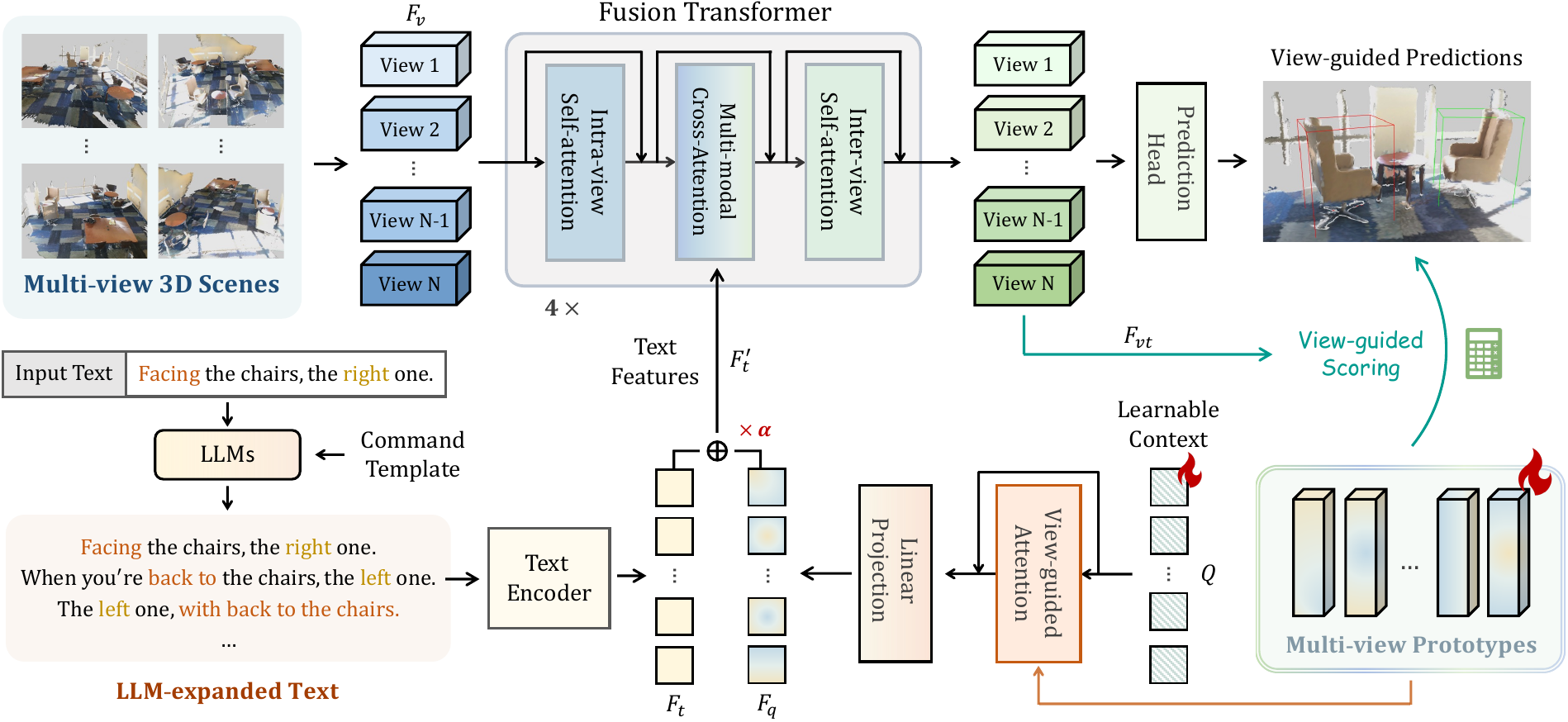}
   \figcaption{\textbf{Overall Pipeline of ViewRefer for 3D Visual Grounding.} We leverage LLMs to enrich the input texts, and introduce a fusion transformer with inter-view attention to enhance cross-view interaction. Upon that, we propose learnable multi-view prototypes to boost the multi-modal fusion via view-guided textual context and scoring strategy. }
    \label{pipeline}
    \vspace{0.1cm}
\end{figure*}

3D Visual Grounding task aims to locate the targeted object in a 3D scene according to a natural language expression. As the baselines, Scanrefer~\cite{chen2020scanrefer} and Referit3D~\cite{achlioptas2020referit3d} first propose datasets for 3D visual grounding that contain object-annotation pairs on ScanNet~\cite{dai2017scannet}. Most recent works~\cite{he2021transrefer3d,yuan2021instancerefer,zhao20213dvg,roh2022languagerefer,yang2021sat,huang2022multi,wu2022eda} adopt a two-stage pipeline that leveraging ground truth or a 3D object detector~\cite{kirillov2019panoptic,jiang2020pointgroup} to obtain the object proposals, utilizing text and 3D encoder~\cite{qi2017pointnet,qi2017pointnet++,zhang2023nearest,zhang2023parameter} to extract features, and then grounding the target one after the feature fusion.
Among them, InstanceRefer~\cite{yuan2021instancerefer} simplifies the task by treating it as an instance-matching problem.
LanguageRefer~\cite{roh2022languagerefer} transforms the multi-modal task into a language modeling problem by replacing the 3D features with predicted object labels.
SAT~\cite{yang2021sat} utilizes extra 2D semantics to enhance the multi-modal alignment.
MVT~\cite{huang2022multi} first attempts to alleviate the view discrepancy issue by starting from 3D modality to build a view-robust multi-modal representation.
Different from the two-stage methods, 3D-SPS~\cite{luo20223d} treats the 3D visual grounding task as a keypoint selection problem, and first proposes a single-stage network to bridge the gap between detection and matching. 
Some of the works above point out the crucial view discrepancy issue in 3D visual grounding and propose some preliminary designs to address it. Unlike prior works that only focus on solving the issue from the 3D modality, we start by grasping view knowledge from both text and 3D modalities to address the challenging view discrepancy problem.

\paragraph{Multi-view Learning in 3D.}
Some recent studies~\cite{su2015multi,brazil2019m3d,shi2019pointrcnn,goyal2021revisiting,hong20233d,hong20223d} in 3D vision concentrate on improving representation learning by generating 2D renderings from 3D under multiple viewpoints. MVCNN~\cite{su2015multi} generates a large number of 2D images, and simply encodes them for 3D shape classification. 
PointCLIP~\cite{zhang2022pointclip} transfers 2D knowledge into the 3D domain via multi-view projection for zero-shot learning.
% by projecting into 2D region proposals and then projecting them back into 3D. 
% PointRCNN~\cite{shi2019pointrcnn} incorporates multi-view representations via a voxel-based fusion method to combine the 3D point cloud from different views. 
SimpleView~\cite{goyal2021revisiting} proposes an effective multi-view framework for shape classification in 3D point cloud learning.
Following the framework of~\cite{zhang2022point}, I2P-MAE~\cite{zhang2022learning} utilizes projected multi-view 2D depth maps to guide 3D point cloud pre-training.
DETR3D~\cite{wang2022detr3d}, PETR~\cite{liu2022petr}, and related methods~\cite{hong2022cross,huang2022tig} conduct 3D object detection based on multi-view images.
These works have shown the effectiveness of multi-view representations in enhancing the performance and robustness of various 3D tasks. For our 3D visual grounding task, we introduce ViewRefer to effectively grasp the multi-view cues from multi-modal input data.

\vspace{0.1cm}
\paragraph{Multi-modality Learning in 3D.}
Multi-modality learning aims at learning from signals of multiple modalities at the same time, which obtains more robust performance than single-modality learning. Recently, multiple multi-modality networks in 2D vision~\cite{radford2021learning,gao2021clip,zhang2022tip,coop,cocoop,guo2022calip,lin2022frozen,zhang2022can,qiu2021vt,zhang2022collaboration,zhang2023prompt,yu2021vector,tan2019lxmert} show effective performance. Inspired by them, PointCLIP V1~\cite{zhang2022pointclip} and V2~\cite{zhu2022pointclip} introduce multi-modality into 3D by adopting CLIP's pre-trained knowledge. What's more,~\cite{zhang2021self,zhang2021dspoint} conduct point-voxel joint-training design, and CrossPoint~\cite{afham2022crosspoint} proposes an image-point contrastive learning network. Also, via filter inflation, Image2point~\cite{xu2021image2point} utilizes pre-trained 2D knowledge for point cloud understanding.
With the emergence of MAE~\cite{he2022masked}, some works~\cite{zhang2022learning,guo2023joint,chen2023pimae} combine multi-modality learning with MAE-based pre-training paradigm and achieve great representation capabilities. 
% Additionally, for 3D object detection, CMKD~\cite{hong2022cross} and TIG-BEV~\cite{huang2022tig} benefit 2D-based detectors with spatial cues provided by 3D LiDAR modality. 
For 3D visual grounding, a nature multi-modality task in 3D, ViewRefer imports specific designs for both single-modal feature extraction and multi-modal fusion for precise cross-modal grounding.

\section{Method}
In this section, we illustrate the details of ViewRefer for 3D visual grounding. We first present our overall pipeline in Section~\ref{overall}. Then, in Section~\ref{llm} and~\ref{fusion}, we respectively elaborate on the proposed designs for text and 3D modalities, i.e., LLM-expanded grounding texts and fusion transformer. Finally, in Section~\ref{proto}, we introduce the multi-view prototypes into our framework.

\vspace{0.1cm}
\subsection{Overall Pipeline}
\label{overall}
\vspace{0.1cm}
The whole framework of ViewRefer is shown in Figure~\ref{pipeline}. 
Given the point cloud of a 3D scene, we first rotate its coordinates as $N$ different views and encode the $N$-view features $F_v \in \mathbb{R}^{N \times K \times D}$ following~\cite{huang2022multi}, where $K$ and $D$ denote object number and feature dimension, respectively. 
Meanwhile, for the input text, we propose to feed it into the pre-trained large-scale language models (LLMs) and obtain $M$ expanded texts with view cues, denoted as $T$ (Section~\ref{llm}). Then, we adopt BERT~\cite{devlin2018bert} as the text encoder to extract the LLM-expanded text features as $F_t \in \mathbb{R}^{M \times L \times D}$, where $L$ denotes the max sequence length. 

On top of that, the $N$-view 3D features and $M$ view-guided text features are fed into fusion transformer with cascaded blocks for multi-modal interactions (Section~\ref{fusion}). Each block sequentially contains the layers for intra-view self-attention, multi-model cross-attention, and inter-view self-attention. By this fusion transformer, the multi-view 3D features can sufficiently interact with each other and incorporate grounding information from the expanded text features. After this, the fused features, denoted as $F_{vt}$, are passed into a prediction head for multi-view grounding logits, which are finally aggregated across views as the output.

To better grasp the latent view cues, we propose a set of multi-view prototypes learnable during training, and leverage them to assist the grounding from two aspects (Section~\ref{proto}). First, we leverage the prototypes to produce view-guided context $F_q$ to the text domain, which is then combined with text features as $F_t'$ before feeding into the fusion transformer. Second, the prototypes are also utilized for weighing the importance of different views. By a view-guided scoring strategy, we further inject multi-view knowledge into the final logits for better grounding results.

We follow previous works~\cite{roh2022languagerefer,huang2022multi} to adopt three losses on ViewRefer, which are cross-entropy loss $L_{ref}$ on the grounding predictions, $L_{text}$ of text classification on the text encoder, and $L_{3D}$ of 3D shape classification upon the object encoder. The whole loss function is formulated as
\begin{align}
\label{loss}
\begin{split}
    & L = L_{ref} + \beta \cdot L_{text} + \gamma \cdot L_{3D},
\end{split}
\end{align}
where $\beta$ and $\gamma$ denote the weights for $L_{text}$ and $L_{3D}$. Both of them are set as $0.5$ in our method.

\vspace{0.3cm}
\subsection{LLM-expanded Grounding Texts}
\label{llm}
To fully exploit view knowledge within the text modality, we propose to utilize the abundant linguistic knowledge in LLMs, e.g., GPT-3~\cite{brown2020language}, to expand the original simple grounding texts.
As discussed in~\cite{achlioptas2020referit3d}, the 3D-text data in 3D visual grounding task can be divided into view-dependent and view-independent pairs, depending on whether the expression is constrained to the speaker’s perspective. For both of them, we construct general command templates fed into the GPT-3 to generate $M$ LLM-expanded texts. Further, considering their different semantic compositions, we adopt specific expanding strategies for the two categories.

\paragraph{View-Dependent Text.}
\label{vdtext}
The description in such texts is strongly related to speaker's viewpoint.
Therefore, for an input view-dependent text, we first generate its synonymous sentence that refers to the same target but under different speaker's views.
Specifically, we create a view-related phrase dictionary, the keys of which include phrases for orientation like ``looking at'', and phrases describing positional relationships like ``left''. The corresponding values to the keys are the opposite phrases of them, e.g., ``looking at - with back to'', and ``left - right''. Then, we construct a command template as the input for GPT-3, which includes the input text and aforementioned view-related phrases in the dictionary. Such general template is designed as
\begin{align}
\begin{split}
% \nonumber
&\operatorname{``\ Rephrase}\ \operatorname{the}\ \operatorname{sentence}\ \operatorname{of}\ \operatorname{`\textcolor{gray}{[TEXT]}'}\\
&\ \ \ \ \ \ \ \ \operatorname{to}\ \operatorname{the}\ \operatorname{opposite}\ \operatorname{perspective,}\\
&\operatorname{which}\ \operatorname{contains}\ \operatorname{phrases}\ \operatorname{`\textcolor{gray}{[PHRASEs]}':\ ''}
\end{split}
\label{template1}
\end{align}
where $\operatorname{\textcolor{gray}{[TEXT]}}$ and $\operatorname{\textcolor{gray}{[PHRASEs]}}$ are replaced by the input text and view-related phrases, respectively. For example, given the initial text of ``\textcolor{gray}{\textit{Facing the front of the couch, pick the table that is to the right of the couch}}'', we complete the command as
\begin{align}
\begin{split}
\nonumber
&\operatorname{``\ Rephrase\ the\ sentence\ of\ `\textcolor{gray}{\textit{Facing\ the\ front\ of}}}\\
&\operatorname{\ \ \ \textcolor{gray}{\textit{the\ couch,\ pick\ the\ table\ that\ is\ to\ the\ right}}}\\
&\operatorname{\ \ \ \ \textcolor{gray}{\textit{of\ the\ couch}}'\ to\ the\ opposite\ perspective,}\\
&\operatorname{\ which\ contains\ phrase\ `\textcolor{gray}{\textit{with\ back\ to}}',\ `\textcolor{gray}{\textit{left}}':\ ''}.
\end{split}
\end{align}
\begin{figure}[t!]
\vspace{-0.3cm}
\centering
    \includegraphics[width=0.97\linewidth]{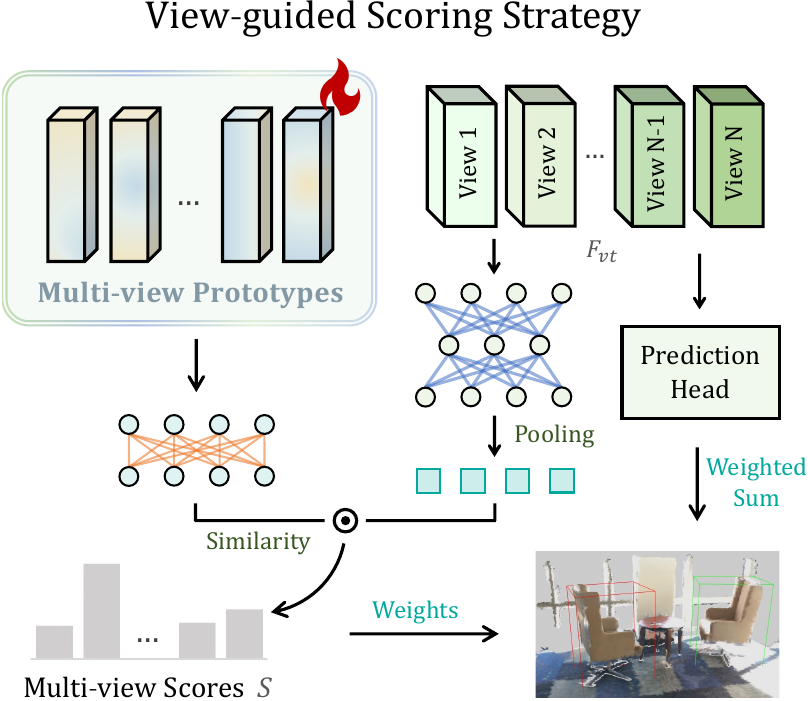}
    \vspace{0.1cm}
  \caption{\textbf{View-guided Scoring Strategy.} Guided by the learnable multi-view prototypes, we assess the importance of each view to predict the final grounding logits.}
  \label{fig_score}
\end{figure}

\noindent By this command, we obtain GPT-3's response as \textit{``With back to the front of the couch, pick the table that is on the left side of the couch"}. In this way,  the rephrased description contains opposite-view information. Together with the original single text, we enrich the text modality with more view-dependent semantics.
To further unleash LLMs' linguistic knowledge, we respectively rephrase the input sentence and its opposite-view synonym for $(M-2)/2$ times with a command template like
\begin{align}
\begin{split}
\label{template}
% \nonumber
&\operatorname{``\ Rephrase}\ \operatorname{the}\ \operatorname{sentence}\ \operatorname{of}\ \operatorname{`\textcolor{gray}{[TEXT]}':\ ''}
\end{split}
\end{align}
where $\operatorname{\textcolor{gray}{[TEXT]}}$ is replaced by the input text or its opposite-view synonym. Additionally, we also provide pre-defined replaced pairs as input prompts for rephrasing, e.g., `near'-`next to'-`beside'. 
Through this expanding strategy, we enrich the initial single text into a set of $M$ LLM-expanded texts, which contain abundant multi-view cues and effectively alleviate the discrepancy issue.

\vspace{0.1cm}
\paragraph{View-Independent Text.}
The description in view-independent texts is irrelevant to speaker's viewpoint.
Thus, we adopt no opposite-view enrichment on these view-independent inputs, and directly rephrase them for $M-1$ times with Template~\ref{template} with pre-defined replaced pairs as input prompts. The generated $M$ LLM-expanded texts also inherit the rich linguistic knowledge from LLMs, and diversify the text embedding space for visual grounding performance.

\begin{figure*}[t!]
\centering
\small
\vspace{-0.1cm}
\centering
	\begin{tabular}{l|ccccc|ccccc}
	\toprule
        \makecell*[c]{\multirow{3}*{Method}} &\multicolumn{5}{c}{Sr3D}\vline &\multicolumn{5}{c}{Nr3D}\\
        \cmidrule(lr){2-6} \cmidrule(lr){7-11} 
	&\makecell*[c]{\multirow{2}*{Overall}} 
        &\makecell*[c]{\multirow{2}*{Easy}} 
        &\makecell*[c]{\multirow{2}*{Hard}} 
        &\makecell*[c]{\multirow{2}*{\shortstack{\vspace*{0.1pt}\\View\\\vspace*{0.2pt}\\Dep.}}} 
        &\makecell*[c]{\multirow{2}*{\shortstack{\vspace*{0.1pt}\\View\\\vspace*{0.2pt}\\Indep.}}} &\makecell*[c]{\multirow{2}*{Overall}} 
        &\makecell*[c]{\multirow{2}*{Easy}} 
        &\makecell*[c]{\multirow{2}*{Hard}} 
        &\makecell*[c]{\multirow{2}*{\shortstack{\vspace*{0.1pt}\\View\\\vspace*{0.2pt}\\Dep.}}} 
        &\makecell*[c]{\multirow{2}*{\shortstack{\vspace*{0.1pt}\\View\\\vspace*{0.2pt}\\Indep.}}}\\
        &&&&&&&&&&\\
  % \makecell*[c]{\multirow{2}*{\shortstack{\vspace*{2.2pt}\\View\\\vspace*{0.3pt}\\Dep.}}}
		 % \cmidrule(lr){1-1} \cmidrule(lr){2-4} \cmidrule(lr){5-6}
         \cmidrule(lr){1-1} \cmidrule(lr){2-2} \cmidrule(lr){3-3} \cmidrule(lr){4-4} \cmidrule(lr){5-5} \cmidrule(lr){6-6} \cmidrule(lr){7-7} \cmidrule(lr){8-8}
         \cmidrule(lr){9-9} \cmidrule(lr){10-10}
         \cmidrule(lr){11-11} 
            ReferIt3D~\cite{achlioptas2020referit3d} &40.8$\%$&44.7$\%$&31.5$\%$&39.2$\%$&40.8$\%$ &35.6$\%$&43.6$\%$&27.9$\%$&32.5$\%$&37.1$\%$\\
            TGNN~\cite{tgnn} &45.0$\%$&48.5$\%$&36.9$\%$&45.8$\%$&45.0$\%$ &37.3$\%$&44.2$\%$&30.6$\%$&35.8$\%$&38.0$\%$\\
            InstanceRefer~\cite{yuan2021instancerefer} &48.0$\%$&51.1$\%$&40.5$\%$&45.4$\%$&48.1$\%$ &38.8$\%$&46.0$\%$&31.8$\%$&34.5$\%$&41.9$\%$\\
            3DVG-Transformer~\cite{zhao20213dvg} &51.4$\%$&54.2$\%$&44.9$\%$&44.6$\%$&51.7$\%$&40.8$\%$&48.5$\%$&34.8$\%$&34.8$\%$&43.7$\%$\\
            LanguageRefer~\cite{roh2022languagerefer} &56.0$\%$&58.9$\%$&49.3$\%$&49.2$\%$&56.3$\%$ &43.9$\%$&51.0$\%$&36.6$\%$&41.7$\%$&45.0$\%$\\
            TransRefer3D~\cite{he2021transrefer3d} &57.4$\%$&60.5$\%$&50.2$\%$&49.9$\%$&57.7$\%$ &42.1$\%$&48.5$\%$&36.0$\%$&36.5$\%$&44.9$\%$\\
            % \textcolor{notecolor}{\CIRCLE} 
            SAT~\cite{yang2021sat} &57.9$\%$&61.2$\%$&50.0$\%$&49.2$\%$&58.3$\%$ &49.2$\%$&56.3$\%$&42.4$\%$&46.9$\%$&50.4$\%$\\
            MVT~\cite{huang2022multi} &64.5$\%$&66.9$\%$&58.8$\%$&58.4\%&64.7$\%$ &55.1$\%$&61.3$\%$&49.1$\%$&54.3$\%$&55.4$\%$\\
            \cmidrule(lr){1-11} 
            MVT$^*$\cite{huang2022multi} &64.2$\%$&67.3$\%$&57.0$\%$&~\textbf{55.6\%}&64.6$\%$ &54.5$\%$&61.2$\%$&48.0$\%$&53.0$\%$&55.2$\%$\\
            \rowcolor{notecolor!8}\textbf{ViewRefer} &~\textbf{67.0\%} &~\textbf{68.9\%} &~\textbf{62.1\%} &~52.2\% &~\textbf{67.7\%} &~\textbf{56.0\%} &~\textbf{63.0\%} &~\textbf{49.7\%} &~\textbf{55.1\%} &~\textbf{56.8\%}\\
            % \textit{Gain} &\textcolor{notecolor2}{+2.8\%} &\textcolor{notecolor2}{+1.6\%} &\textcolor{notecolor2}{+5.1\%} &\textcolor{notecolor2}{+1.3\%} &\textcolor{notecolor2}{+3.2\%} &\textcolor{notecolor2}{+1.6\%} &\textcolor{notecolor2}{+1.9\%} &\textcolor{notecolor2}{+2.0\%} &\textcolor{notecolor2}{+2.1\%} &\textcolor{notecolor2}{+1.7\%} \\
	\bottomrule
	\end{tabular}
 \tabcaption{\textbf{Performance of ViewRefer on Sr3D and Nr3D}. We report the performance on the overall dataset and all its splits. `*' denotes our implementation results$^\dagger$.}
\vspace{-0.4cm}
 \label{sr3d}
\end{figure*}

\vspace{0.3cm}
\subsection{Fusion Transformer}
\vspace{0.1cm}
\label{fusion}

% \begin{figure*}[t!]
% \vspace{-0.2cm}
%     \includegraphics[width=0.97\textwidth]{./gpt3.pdf}
%    \caption{\textbf{Illustration of the Input 3D Scenes, Initial Texts and LLM-expanded texts.} In each column, we show the corresponding 3D-text pairs with their LLM-expanded texts. In the utterances, the words in orange are phrases for orientation, in yellow are phrases describing the positional relationship, and the green ones are the target objects. The first two lines in the LLM-expanded texts are synonymous sentences with opposite views, while the last lines are from the same perspectives.}
%     \label{gpt_vis}
% \vspace{-0.2cm}
% \end{figure*}

To integrate 3D multi-view and textual features for grounding, we introduce a multi-modal transformer composed of cascaded fusion blocks. Each block includes three types of attention mechanisms with residual connections. Firstly, an intra-view attention layer is adopted for the $N$-view features $F_v$, which independently interacts $K$ object features within each view. Secondly, we utilize a multi-modal cross-attention layer referring to~\cite{huang2022multi} to infuse textual information from $F'_t$ into the multi-view features. We formulate them as
\begin{align}
\label{2attn}
\begin{split}
F_v &= F_v\  + \operatorname{Intra\text{-}Attn}(F_v),\\
F_{vt} &= F_v + \operatorname{Cross\text{-}Attn}(F_v,\ F'_t),\\
\end{split}
\end{align}
where $F_{vt} \in \mathbb{R}^{N \times K \times D}$ denotes the multi-modal features. %, and $K,D,L$ denote object number, feature dimension, and max sequence length, respectively. 

After that, to boost the cross-view communication, we 
% we conduct a transpose operation on $F_f$ and 
subsequently introduce an inter-view attention layer. Orthogonal to intra-view attention, this layer calculates attention weights across different views for the same object, formulated as
\begin{align}
\label{interview_attn}
\begin{split}
F_{vt} = \left (F_{vt}^T+ \operatorname{Inter\text{-}Attn}(F_{vt}^T)\right )^T.\\
\end{split}
\end{align}
With such interaction, the network can better capture the inter-view differences from one object, and focus on more informative views to encode multi-modal features $F_{vt}$. 

% model can obtain a comprehensive understanding of the spatial relations of the whole 3D scene, which benefits the object grounding effectively. Note that we adopt a transpose operation on $F_f$ before feeding them into the inter-view attention, i.e., 
% \begin{align}
% \label{interview_attn}
% \begin{split}
% &F'_f = \operatorname{Transpose}(F_f),\\
% &F\ \ = F'_f+ \operatorname{Inter\text{-}Attn}(F'_f),\\
% \end{split}
% \end{align}
% where $F\in \mathbb{R}^{N \times K \times D}$ denotes the output fusion features.

\vspace{0.3cm}
\subsection{Multi-view Prototypes}
\label{proto}
Besides our LLM-expanded texts and fusion transformer, we further introduce a set of multi-view prototypes to conduct view-guided 3D visual grounding in ViewRefer. 
The prototypes are randomly initialized embeddings and represented as $Proto \in \mathbb{R}^{N \times D}$, where $N,D$ denote the view number and feature dimension. They aim to memorize the 3D-text view-consistent knowledge from a common multi-modal space, which serve as high-level guidance for the grounding process from the following two aspects. 

% We respectively 

% One is a view-guided attention module adopted on the text features, which incorporates the multi-modal aligned view knowledge to boost textual semantics. 

% The other is the view-guided scoring strategy, which utilizes the 3D-text view-consistency knowledge of multi-view prototypes to assess the relative importance of each view, for precise view-guided visual grounding. In the next two paragraphs, we elaborate on the details of them.

% \blfootnote{$^\dagger$We implement the results based on the open-source code, with a single NVIDIA A100 GPU.}

\paragraph{View-guided Textual Context.}
\label{vg_attn}
We first utilize the prototypes to incorporate view-guided contexts into the textual features $F_t$.
% introduce a view-guided attention module to refine the features of multiple LLM-expanded texts via 3D-text view-consistency knowledge from learned multi-view prototypes. 
As shown in the bottom right of Figure~\ref{pipeline}, we first randomly initialize a learnable context $Q\in \mathbb{R}^{M \times D}$, which serves as `query' to extract 3D-text multi-view information from the prototypes $Proto$. 
Then, we regard the prototypes as `key' and `value', and propose a view-guided attention module with residual connections for feature interaction. We utilize one cross-attention layer in the module, formulated as
% , and get $F_q$ that contains the view-guided context $F_q$ absorbed from $MV\text{-}P$, i.e.,
\vspace{-0.1cm}
\begin{align}
\label{query}
\begin{split}
& F_q = Q + \operatorname{VG\text{-}Attn}(Proto,\ Q),
\end{split}
\end{align}
where $\operatorname{VG\text{-}Attn}$ denotes the view-guided attention module. After this, the query features $F_q$ have adaptively aggregate informative multi-view cues and are element-wisely added to the input text features $F_t$ as
\vspace{-0.1cm}
\begin{align}
\label{res}
\begin{split}
& F'_t = F_t + \alpha \cdot F_q,
\end{split}
\end{align}
where $\alpha$ denotes a frozen balance factor weighing the importance of view-guided context, which is initialized before training. From the interaction with prototypes, the text features $F'_t$ become view-aware and perform better multi-modal fusion in the subsequent transformer.

\begin{figure}[t!]
\vspace{0.3cm}
\centering
\small
\centering
\begin{adjustbox}{width=\linewidth}
	\begin{tabular}{lc c c c c}
	\toprule
		\makecell*[c]{\multirow{2}*{Method}} &\multicolumn{2}{c}{Multiple} &\multicolumn{2}{c}{Overall}\\
		 \cmidrule(lr){2-3} \cmidrule(lr){4-5} 
		 &Acc.@0.25 &Acc.@0.50 &Acc.@0.25 &Acc.@0.50 \\
		 \cmidrule(lr){1-1} \cmidrule(lr){2-5}
            ReferIt3D~\cite{achlioptas2020referit3d}  &21.03$\%$ &12.83$\%$ &26.44$\%$ &16.90$\%$ \\
		    ScanRefer~\cite{chen2020scanrefer} &30.63$\%$ &19.75$\%$ &37.30$\%$ &24.32$\%$ \\
            TGNN~\cite{tgnn} &27.01$\%$ &21.88$\%$ &34.29$\%$ &27.92$\%$ \\
            InstanceRefer~\cite{yuan2021instancerefer} &31.27$\%$ &24.77$\%$ &40.23$\%$ &32.93$\%$ \\
            MVT~\cite{huang2022multi} &31.92$\%$ &25.26$\%$ &40.80$\%$ &33.26$\%$ \\
            \cmidrule(lr){1-5} 
            MVT$^*$~\cite{huang2022multi} &31.46$\%$ &24.85$\%$ &39.95$\%$ &32.28$\%$ \\
            \rowcolor{notecolor!5}\textbf{ViewRefer} &~\textbf{33.08\%} &~\textbf{26.50\%} &~\textbf{41.30\%} &~\textbf{33.66\%}\\
            % \textit{Gain} &\textcolor{notecolor2}{+1.62\%} &\textcolor{notecolor2}{+1.65\%} &\textcolor{notecolor2}{+1.35\%}&\textcolor{notecolor2}{+1.38\%} \\
	\bottomrule
	\end{tabular}
 \end{adjustbox}
\tabcaption{\textbf{Performance of ViewRefer on ScanRefer}. We report the Acc@0.25 and Acc@0.50 on ScanRefer dataset. `*' denotes our implementation results$^\dagger$.}
 \label{scanrefer}
\end{figure}

\paragraph{View-guided Scoring Strategy.}
\label{score}
Furthermore, we leverage the multi-view prototypes to conduct a scoring for the predicted logits after the fusion transformer. As shown in Figure~\ref{fig_score}, we calculate $N$-view scores for the multi-modal features $F_{vt}$, which assess the relative importance of each view in the final prediction process.
We first apply a pooling operation to $F_{vt}$ along the dimension of intra-view objects, which aggregates the global features of different views. We then adopt linear layers to project the prototypes and multi-modal features into a unified embedding space.
Upon that, we calculate the multi-view scores by matrix multiplication of each $N$-view pair as
\vspace{-0.1cm}
\begin{align}
\begin{split}
& S = \operatorname{MatMul}\big(Proto,\ \operatorname{Pooling}(F_{vt})\big),
\end{split}
\end{align}
where $S\in \mathbb{R}^{N \times 1}$ denotes the score for each $N$ view. After this, the multi-view scores serve as the weights to aggregate predicted logits among different views. Such scoring can adaptively constrain the prediction of those views inconsistent with the input text, and emphasize the informative views for accurate 3D visual grounding.

\vspace{0.2cm}
\section{Experiments}
\label{exp}
In this section, we evaluate the performance of ViewRefer on three commonly used benchmarks, i.e., Sr3D~\cite{achlioptas2020referit3d}, Nr3D~\cite{achlioptas2020referit3d}, and ScanRefer~\cite{chen2020scanrefer}.

\vspace{0.4cm}
\subsection{Datasets}
\vspace{0.2cm}
\paragraph{Sr3D and Nr3D.}
The Sr3D dataset~\cite{achlioptas2020referit3d} contains 83,572 template-based utterances leveraging spatial relationships among objects to localize a referred object in a 3D scene. The scenes are constrained to have no more than six distractors, i.e., objects belonging to the same category as the target. 
The Nr3D dataset~\cite{achlioptas2020referit3d} provides annotations for the ScanNet~\cite{dai2017scannet}, an indoor 3D scene dataset, comprising 45,503 natural and free-form utterances. The dataset includes 707 distinct indoor scenes with target objects belonging to 76 fine-grained categories.
Two distinct data splits are employed in Sr3D and Nr3D, namely the ``Easy" and ``Hard" splits that differ based on the number of distractors in the scene, and the ``View-dependent" and ``View-independent" splits that vary based on whether the referring expression relies on the speaker's viewpoint.

\paragraph{ScanRefer.}
\blfootnote{$^\dagger$We reproduce the results based on the official open-source code~\cite{huang2022multi} on a single NVIDIA A100 GPU.}
The ScanRefer dataset~\cite{chen2020scanrefer} annotates 800 indoor scenes in ScanNet~\cite{dai2017scannet} and contains 51,583 utterances, composed of 36,665 samples for the training set, 9,508 for val set, and 5,410 for the test set.
% Depending on whether there are objects of
% the same target class in the scene, the dataset can be split into two parts, i.e., the ``Unique" part and the ``Multiple" one.

\vspace{0.4cm}
\subsection{Experimental Settings}
\label{setting}
\paragraph{Evaluation Metrics.} 
Following existing works~\cite{huang2022multi,yuan2021instancerefer}, we utilize the ground truth object proposals for Nr3D and Sr3D~\cite{achlioptas2020referit3d} datasets, while adopting detector-generated object proposals via pre-trained detector, PointGroup~\cite{jiang2020pointgroup}, for ScanRefer~\cite{chen2020scanrefer} dataset. As for the evaluation metrics, we follow previous works~\cite{huang2022multi,yuan2021instancerefer} to measure the networks via the accuracy of the target predictions for Nr3D and Sr3D, and the Acc@$m$IoU metric for ScanRefer where, $m \in \{0.25, 0.50\}$. The Acc@$m$IoU measures the proportion of text input with the predicted bounding box overlapping the ground truth box by intersection over the union (IoU) higher than $m$.

\paragraph{Implementation Details.} 
\begin{figure*}[t!]
\vspace{-0.2cm}
    \includegraphics[width=\textwidth]{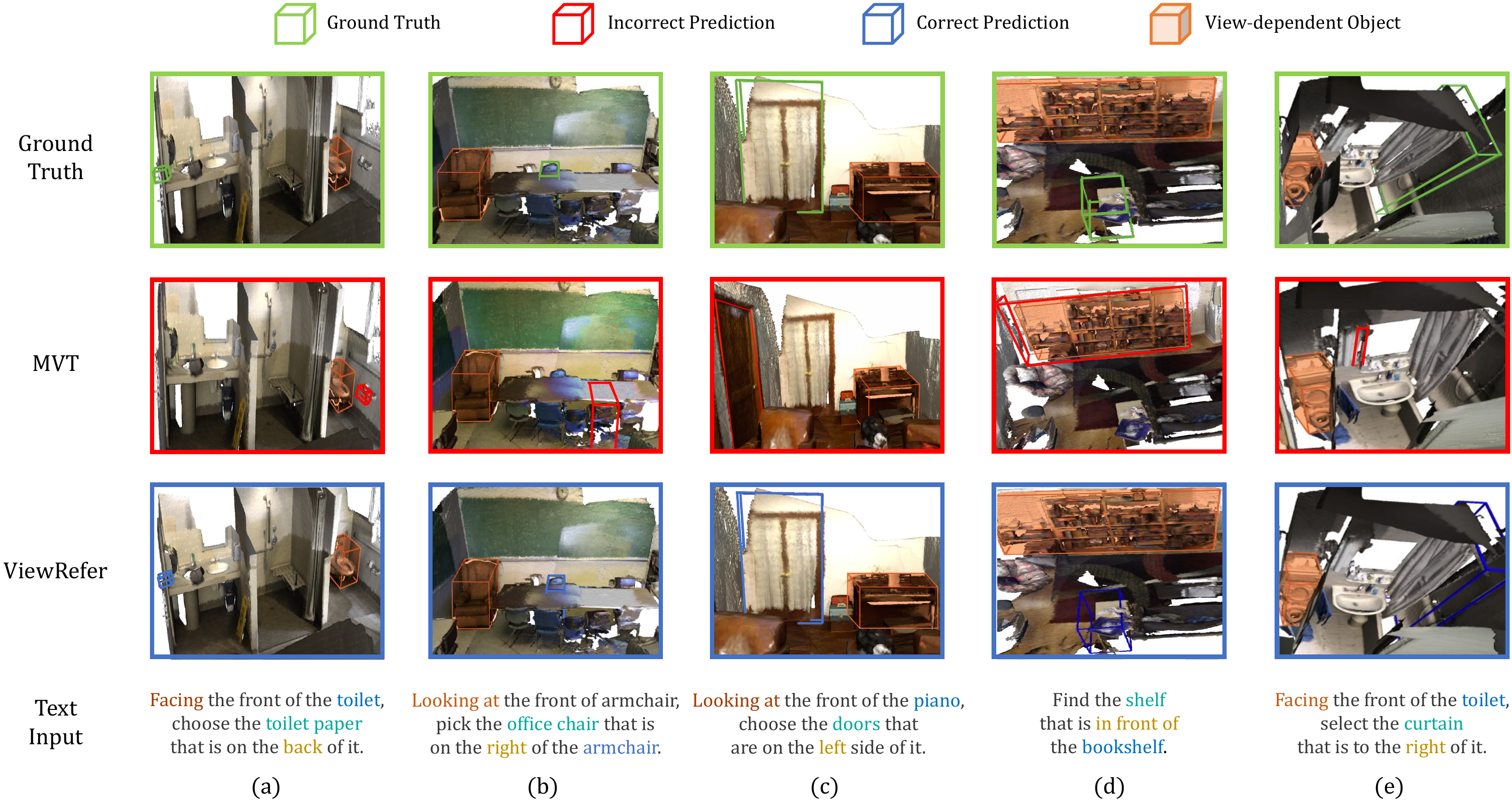}
   \caption{\textbf{Visualization of the 3D Visual Grounding Results.} For the presented 3D scenes, we utilize green, red, blue, and orange boxes to represent the ground truth, incorrect predictions, correct predictions, and view-dependent objects, respectively.}
    \label{res_vis}
\end{figure*}

\begin{figure*}[t!]
\centering
\small
\centering
	\begin{tabular}{cc|c c|c c|c c c c c}
	\toprule
		\makecell*[c]{\multirow{2}*{Decoder}} &\makecell*[c]{\multirow{2}*{\shortstack{\vspace*{2.2pt}\\Multi-view\\\vspace*{0.3pt}\\Input}}} &\makecell*[c]{\multirow{2}*{\shortstack{\vspace*{2.2pt}\\Inter-view\\\vspace*{0.3pt}\\Attention}}} &\makecell*[c]{\multirow{2}*{\shortstack{\vspace*{2.2pt}\\LLM-expanded \\\vspace*{0.3pt}\\Text}}} &\multicolumn{2}{c}{Multi-view Prototype}\vline &\makecell*[c]{\multirow{2}*{Overall}} &\makecell*[c]{\multirow{2}*{Easy}} &\makecell*[c]{\multirow{2}*{Hard}} &\makecell*[c]{\multirow{2}*{\shortstack{\vspace*{2.2pt}\\View\\\vspace*{0.3pt}\\Dep.}}} &\makecell*[c]{\multirow{2}*{\shortstack{\vspace*{2.2pt}\\View\\\vspace*{0.3pt}\\Indep.}}}\\\cmidrule(lr){5-6}
		 &&& &VG. Score. & VG. Cont. &\\
		 \cmidrule(lr){1-1} \cmidrule(lr){2-2} \cmidrule(lr){3-3} \cmidrule(lr){4-4} \cmidrule(lr){5-5} \cmidrule(lr){6-6} \cmidrule(lr){7-7} \cmidrule(lr){8-8} \cmidrule(lr){9-9} \cmidrule(lr){10-10} \cmidrule(lr){11-11} 
		\rowcolor{gray!6}-&-&-&-&-&-&22.4$\%$&23.9$\%$&18.7$\%$&26.9$\%$&22.2$\%$\\
        \rowcolor{gray!6}\checkmark&-&-&-&-&-&61.5$\%$&64.2$\%$&55.2$\%$&50.6$\%$&61.9$\%$\\
        % \rowcolor{gray!12}
        \checkmark&\checkmark&-&-&-&-&64.4$\%$&67.0$\%$&58.6$\%$&51.5$\%$&65.0$\%$\\
        % \cmidrule(lr){1-7}
        \checkmark&\checkmark&\checkmark&-&-&-&65.0$\%$&67.2$\%$&59.9$\%$&51.7$\%$&65.6$\%$\\
        \checkmark&\checkmark&\checkmark&\checkmark&-&-&66.0$\%$&68.4$\%$&60.4$\%$&51.9$\%$&66.4$\%$\\
        \checkmark&\checkmark&\checkmark&\checkmark&\checkmark&-&66.5$\%$&68.7$\%$&60.8$\%$&52.1$\%$&66.9$\%$\\
        \rowcolor{notecolor!7}\checkmark&\checkmark&\checkmark&\checkmark&\checkmark&\checkmark&\textbf{67.0$\%$}&\textbf{68.9$\%$}&\textbf{62.1$\%$}&\textbf{52.2$\%$}&\textbf{67.7$\%$}\\
	\bottomrule
	\end{tabular}
\tabcaption{\textbf{Ablation Study on Different Components for Grasping Multi-view Knowledge on Sr3D.} {`VG. Score.' and `VG. Cont.'} denote the view-guided scoring strategy and view-guided textual context, respectively.}
\label{component}
% \end{adjustbox}
\end{figure*}

For Sr3D and Nr3D datasets~\cite{achlioptas2020referit3d}, we train the network for 100 epochs with a batch size of $24$. We utilize AdamW~\cite{kingma2014adam} as the optimizer with an initial learning rate of 5$\times$10$^{-5}$ for the fusion transformer, 5$\times$10$^{-4}$ for other modules, and weight decay as 1$\times$10$^{-3}$. After 40 epochs, we decline the learning rate by multiplying 0.65 per 10 epochs. For ScanRefer~\cite{chen2020scanrefer}, we set the max epoch number as 30, and batch size as 32. We adopt AdamW~\cite{kingma2014adam} as the optimizer with an initial learning rate of 5$\times$10$^{-4}$  with no weight decay.
As for the structure of the network, we set both the view number $N$ and the generated text number $M$ as 4. We utilize 3 layers for the text encoder and 4 transformer blocks for the fusion transformer with 8 attention heads. For data augmentation, on Sr3D and Nr3D, we conduct translation on each 3D object and random rotation on 3D scenes during training. Additionally, we adopt the flip operation on Sr3D dataset. We utilize the same data augmentation with previous works~\cite{huang2022multi,yuan2021instancerefer} on ScanRefer dataset. As for the pooling operation in the view-guided scoring strategy, we adopt sum pooling for Sr3D and ScanRefer datasets, and the summation of max and average pooling for Nr3D.

\vspace{0.7cm}
\subsection{Quantitative Analysis}
\paragraph{Performance on Sr3D.}
In Table~\ref{sr3d}, we report the performance of ViewRefer on Sr3D dataset~\cite{achlioptas2020referit3d} for 3D visual grounding. Our ViewRefer outperforms prior works in all splits, surpassing the \textit{state-of-the-art} method~\cite{huang2022multi} on the overall accuracy by +2.8\%, which demonstrates the effectiveness of the view knowledge grasped by ViewRefer.
Also, the significant amplification of +5.1\% on the challenging ``Hard" split shows the effectiveness of our ViewRefer in 3D scenes with complex spatial relations between objects, which further indicates our superior understanding of spatial relations.

\begin{figure*}[t!]
\vspace{-0.2cm}
  \centering
    \includegraphics[width=0.95\textwidth]{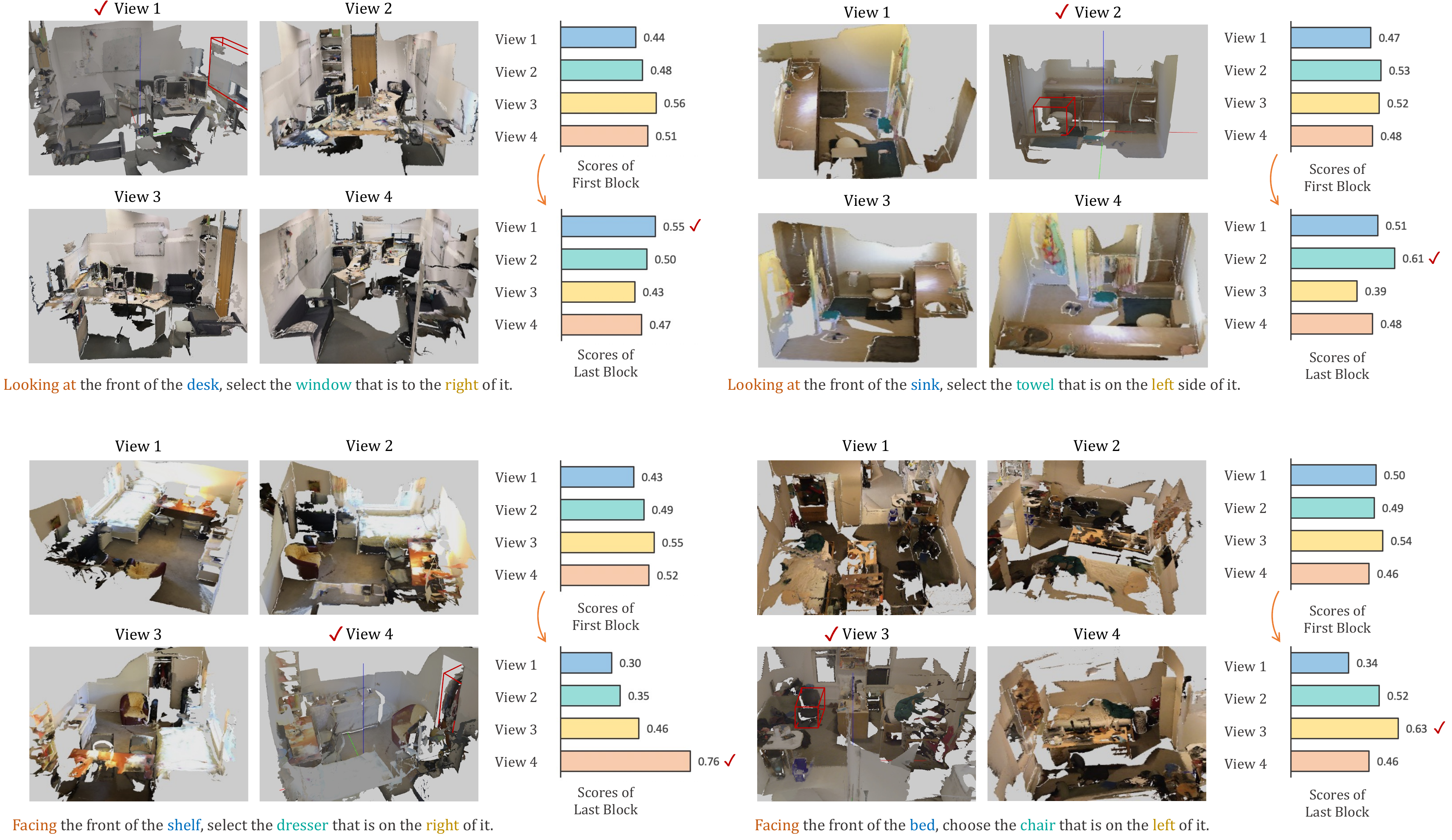}
   \caption{\textbf{Visualization of View-guided Scoring Strategy.} We present four view-dependent samples, and illustrate the multi-view scores calculated from the first to the last fusion transformer blocks. We mark the correct view with `\textcolor{linkcolor}{$\checkmark$}'.}
    \label{score_vis}
    \vspace{-0.1cm}
\end{figure*}

\paragraph{Performance on Nr3D.} 
In Table~\ref{sr3d}, we evaluate ViewRefer on Nr3D dataset, where ViewRefer exceeds best competitor~\cite{huang2022multi} by an overall improvement of +1.5\%, and +2.1\% on ``view-dependent" split, indicating the superiority of ViewRefer on data with intricate view-dependent natural referential utterances, which further demonstrates the effectiveness of our view-guided designs.

\vspace{-0.1cm}
\paragraph{Results on ScanRefer.}
We also evaluate ViewRefer on ScanRefer~\cite{chen2020scanrefer} and report the Acc@$m$IoU performance of ViewRefer in Table~\ref{scanrefer}. Clearly, compared with prior works, our ViewRefer achieves the highest scores on all metrics. This well illustrates the superior view-understanding capability of our multi-view training framework.

\vspace{0.3cm}
\subsection{Qualitative Analysis}
\vspace{0.2cm}
\paragraph{3D Visual Grounding Results.}
In Figure~\ref{res_vis}, we select some view-dependent cases from Sr3D~\cite{achlioptas2020referit3d} and visualize the grounding truth boxes, predictions of MVT~\cite{huang2022multi}, and predictions of ViewRefer in each column from top to bottom. We emphasize the view-dependent objects in orange boxes with highlights for view alignment.
As shown in (a), MVT fails to ensure the correct view as the text input describes, and grounds to wrong object with the same category, while ViewRefer successfully predicts the target under the correct perspective. From the predictions in (b) and (c), we find although ViewRefer and MVT both understand the correct views, only ViewRefer grounds the right target, which shows that ViewRefer achieves a comprehensive understanding of the spatial relations between the objects. Additionally, in (d) and (e), we observe that ViewRefer shows better object recognition ability, while MVT is relatively easy to predict objects of the wrong categories.

% \begin{table}[t!]
% \vspace{0.1cm}
% \centering
% \small
% \begin{tabular}{cccccc}
%     \toprule
%     \multicolumn{3}{c}{Intra-view Aggregation} &\multicolumn{2}{c}{Context} &\makecell*[c]{\multirow{2}*{\ \ \ Overall\ \ \ }} \\
%     \cmidrule(lr){1-3} \cmidrule(lr){4-5}
%       \ \ Avg\ \  &\ \ Max\ \  &Max+Avg &Each &Glo. &\\
%      \cmidrule(lr){1-1}  \cmidrule(lr){2-2}  \cmidrule(lr){3-3}  \cmidrule(lr){4-4}  \cmidrule(lr){5-5}  \cmidrule(lr){6-6} 
%      \checkmark &-&- &\checkmark &- &64.9\%\\
%      \checkmark &-&- &-&\checkmark  &65.4\%\\
%      - &-&\checkmark &\checkmark &- &66.1\% \\
%      - &-&\checkmark &-&\checkmark &66.2\%\\
%      - &\checkmark&- &\checkmark &- &66.6\% \\
%      \rowcolor{notecolor!5} - &\checkmark&- &- &\checkmark &~\textbf{67.2\%} \\
%     \bottomrule
% \end{tabular}
% \tabcaption{\textbf{Ablation Study on Multi-view Prototypes,} including View-guided Scoring Strategy and Textual Query Context. `Avg', `Max', and `Max+Avg' denote average pooling, max pooling, and the summation of both to obtain the global view features. `Context' denotes the combination of textual query features.}
% \label{proto_abla}
% \end{table}

\vspace{-0.1cm}
\paragraph{Interpretability of View-guided Scoring Strategy.}
To demonstrate the interpretability of the proposed view-guided scoring strategy, we utilize the multi-view prototypes and output features from the fusion transformer blocks, to calculate and analyze the intermediary multi-view scores. 
In Figure~\ref{score_vis}, we present the scores calculated from the first and last blocks of view-dependent cases, where we mark the correct view with ``\textcolor{linkcolor}{$\checkmark$}".
As shown from the last blocks' scores, with the proposed scoring strategy, ViewRefer can effectively assess the relative importance of each view for exact visual grounding. Also, we observe the upward trend of the scores of correct view, which shows that with the view-guided designs, our network gradually captures the view knowledge as the network deepens.

\begin{figure*}[t!]
\vspace{-0.2cm}
\centering
\includegraphics[width=\textwidth]{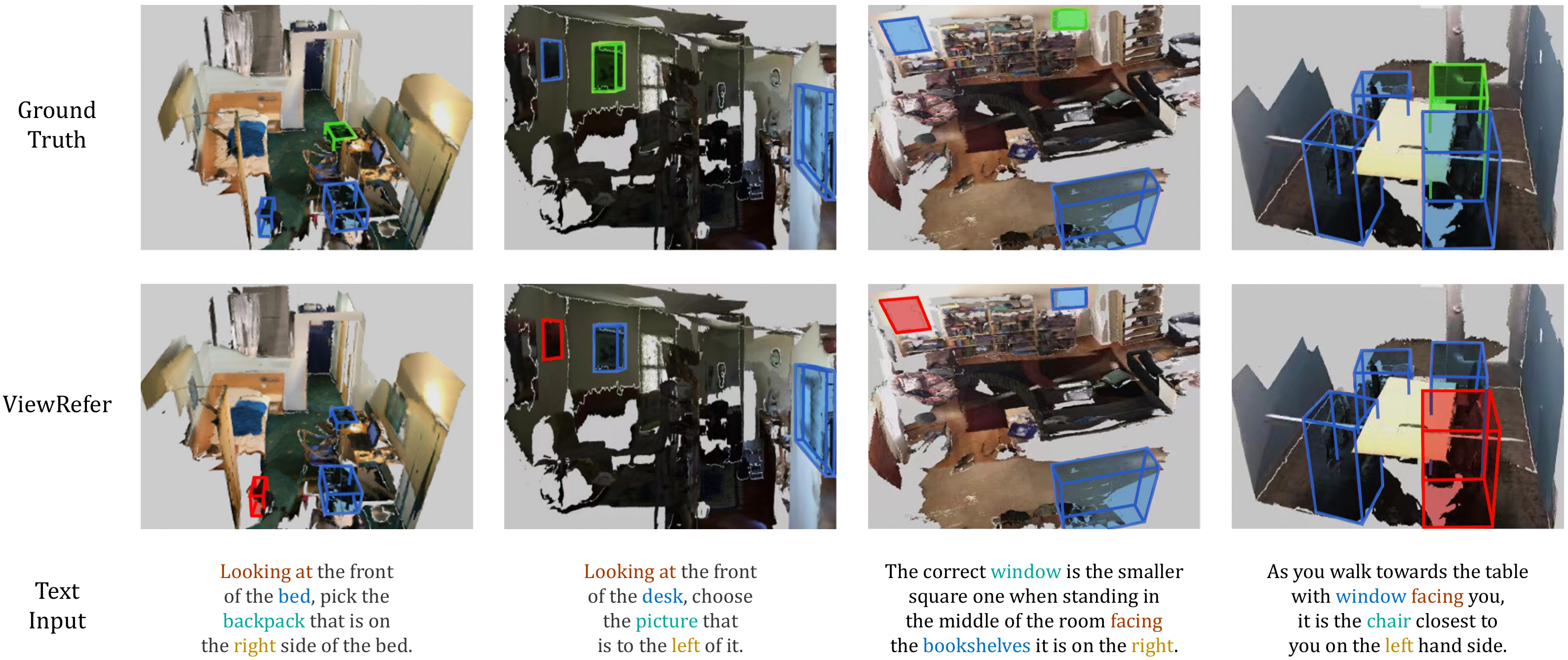}
   \caption{\textbf{Visualization of Failure Cases.} The green boxes and the red boxes represent the ground truth objects and the predictions of ViewRefer, respectively. The blue boxes are the distractors.}
   % In each column, we present the grounding text, the ground truth object, and the predicted box of ViewRefer. 
    \label{failurecase}
\vspace{-0.2cm}
\end{figure*}

\paragraph{Failure Cases.}
In Figure~\ref{failurecase}, we visualize the failure cases of our ViewRefer. We observe that the failure cases can be grouped into two parts. 
One part, like the first two columns, contains overmuch distractors that have the same class as the target object, which critically confuses the model. Also, as shown in the last two columns, the other part's difficulty is the long indigestible ground texts containing too complex grammatical structures for the model to grasp the semantic information. In the future, we will focus on the solutions for these kinds of cases.

\section{Ablation Study}
To explore the effectiveness of each design in ViewRefer, we conduct extensive ablation study on Sr3D~\cite{achlioptas2020referit3d} dataset and evaluate the overall accuracy on 3D visual grounding. In Table~\ref{component}, we report the results of different combinations of proposed designs. In Table~\ref{attn_view}, we explore the positions of inter-view attention and different view numbers.

\begin{table}[t!]
% \begin{adjustbox}{width=0.97\linewidth}
\centering
\small
	\begin{tabular}{ccccc}
    \toprule
    \multicolumn{3}{c}{\ \ \ \ \ \ \ \ \ Position of Inter-view Attn.\ \ \ \ \ \ \ \ \ } &\makecell*[c]{\multirow{2}*{\shortstack{\vspace*{2.2pt}\\View\\\vspace*{0.3pt}\\Number}}} &\makecell*[c]{\multirow{2}*{Overall}} \\
    \cmidrule(lr){1-3}
      \ \ \ \ Before\ \ \ \  &\ \ \ Between\ \ \  &\ \ After\ \ \  & &\\
     \cmidrule(lr){1-1}  \cmidrule(lr){2-2}  \cmidrule(lr){3-3}  \cmidrule(lr){4-4}  \cmidrule(lr){5-5}  
     \checkmark &- &- &4 &65.6\% \\
     - &\checkmark &- &4 &66.6\% \\
     \rowcolor{notecolor!5} - &- &\checkmark &4 &\textbf{67.0\%} \\
     % - &- &\checkmark &1 &64.2\% \\
     - &- &\checkmark &2 &65.3\% \\
     - &- &\checkmark &8 &66.3\% \\
    \bottomrule
\end{tabular}
% \end{adjustbox}
\tabcaption{\textbf{Ablation Study on Inter-view Attention and View Number.} The `Before', `Between', `After' denote block structure of putting the ``inter-view attention" layer before, between, and after the ``intra-view attention" layer and ``multi-modal cross-attention" layer, respectively.}
\label{attn_view}
\vspace{-0.4cm}
\end{table}

\paragraph{Effectiveness of Each Component.}
Based on the baseline network, we conduct ablation studies by adding the designed modules one by one until the final structure of ViewRefer.
In Table~\ref{component}, the first two rows in light gray are baselines with basic fusion modules as described in~\cite{huang2022multi}.
As reported, the import of each major component benefits the object location separately.
% , which shows the effectiveness of each design in ViewRefer.

\paragraph{Fusion Transformer and View Number.}
In Table~\ref{attn_view}, we conduct ablation studies on the structures of fusion transformer and 3D view number. For the fusion transformer, we evaluate ViewRefer with three different designs within the fusion block, which are placing the ``inter-view attention" layer before, between, and after the ``intra-view attention" layer and ``multi-modal cross-attention" layer, respectively. We also explore the influence of 3D view number in Table~\ref{attn_view}. As shown, the fusion transformer with inter-view attention on the last position, and the view number of 4 perform the best. This is because, after the internal interaction among objects within each view, and the fusion between 3D and text, the 3D feature with textual and inner spatial information can boost the cross-view interaction in the inter-view attention layer.
% , which further enhances the visual grounding performance.
Note that subjected to the requirement of frozen view number in the view-guided scoring strategy, we do not evaluate ViewRefer under the setting of adopting different view numbers for training and test phases.

% \paragraph{Implementation of Multi-view Prototypes.}
% Additionally, we conduct ablation study on the implementation regarding the two enhancements of multi-view prototypes, i.e., aggregation in view-guided scoring, and combination of textual query features $F_q$.
% For the aggregation in view-guided scoring, we investigate different pooling operations to integrate features of all objects for a certain view. Meanwhile, we also evaluate two combination types of textual query features: only add the textual query features into the first global token that contains the whole sentence's feature (`Glo.'), or broadcast and add into each token simultaneously (`Each').
% As reported in Table~\ref{proto_abla}, `max pooling' with `global' combination performs the best, which is our default in all experiments. 

\section{Conclusion}
We propose \textbf{ViewRefer}, a multi-view framework for
3D visual grounding that grasps view knowledge from both text and 3D modalities to alleviate the challenging view discrepancy issue.
In the text modality, we introduce LLM-expanded grounding texts to utilize the diverse linguistic knowledge of large-scale language models for view understanding. For 3D modality, we propose a fusion transformer with inter-view attention to grasp rich multi-view knowledge.
Furthermore, with a set of learnable multi-view prototypes, we enhance ViewRefer from two perspectives, view-guided textual context and a view-guided scoring strategy, which further enhances view-guided multi-modal fusion for exact grounding.
Extensive experiments are conducted to demonstrate the superiority of ViewRefer on 3D visual grounding. We expect ViewRefer can inspire more works to further explore the possibility of grasping view knowledge to benefit view understanding in 3D visual grounding.

\paragraph{Acknowledgements}
This work is partially supported by the Shanghai AI Laboratory, National Key R\&D Program of China  (2022ZD0160100) and the National Natural Science Foundation of China (62106183).
% \paragraph{Limitations.}
% Although ViewRefer has verified the effectiveness of exploration of view knowledge from both text and 3D modalities, we will further explore how to fully release the potential abilities of large-scale language models to help 3D visual grounding as the future work. We do not foresee a negative social impact from the proposed work.

\clearpage

{\small
\bibliographystyle{ieee_fullname}
\bibliography{egbib_new}
}

\end{document}